\documentclass[runningheads]{llncs}

\usepackage{eccv}

\usepackage{eccvabbrv}

\usepackage{graphicx}
\usepackage{booktabs}

\usepackage[accsupp]{axessibility}  

\usepackage{hyperref}

\usepackage{orcidlink}
\usepackage{graphicx}
\usepackage{amsmath,amssymb}
\usepackage{booktabs}
\usepackage{xcolor}
\usepackage{colortbl}
\usepackage{multirow}
\begin{document}
\newcommand{\dropcindex}{XX}        
\newcommand{\mmdvisual}{X.XX}      
\newcommand{\mmdvqa}{X.XX}         
\newcommand{\mmdreduction}{YY}     
\newcommand{\numscenarios}{four}
\newcommand{\improvavg}{X.X}       
\title{Semantic-Anchored Evidential Fusion for Domain-Robust Whole-Slide Survival Analysis} 

\titlerunning{Semantic-Anchored Evidential Fusion for WSI Survival}

\author{
Yucheng~Xing\inst{1,2}
\and
Ling~Huang\inst{1,3}\textsuperscript{\ensuremath{\dagger}}
\and
Pei~Liu\inst{4}
\and
Jingying~Ma\inst{1}
\and
Jiaxing~Xu\inst{1}
\and
Kai~He\inst{1}
\and
Mengling~Feng\inst{1}
}

\authorrunning{Y.~Xing et al.}

\institute{
National University of Singapore, Singapore
\and
National University of Singapore Guangzhou Research Translation and
Innovation Institute, Guangzhou, Guangdong, China
\and
Imperial College London, London, UK
\and
Hunan University, Changsha, China
}

\maketitle

\begingroup
\renewcommand{\thefootnote}{\ensuremath{\dagger}}
\footnotetext[0]{Corresponding author: Ling Huang
\textless\href{mailto:my@linghuang.co.uk}
{\texttt{my@linghuang.co.uk}}\textgreater.}
\endgroup

\begin{abstract}
Whole-slide images (WSIs) are widely used for computational cancer prognosis. However, current research primarily focuses on in-domain performance and fails to generalize across clinical centers. This limitation stems from their reliance on pixel-derived representations that are highly susceptible to domain-specific artifacts caused by staining protocols and scanner hardware.
We hypothesize that high-level pathology semantics, such as tumor grade and micro-environmental architecture, provide a domain-invariant semantic representation that mirrors the robust diagnostic logic of human pathologists. Therefore, we propose a Semantic-Anchored Evidential Fusion Survival (SAEFS) framework, where SAEFS derives semantic anchors from WSIs via Visual Question Answering (VQA), employs a dual-stream WSI evidence extraction architecture, uses Dirichlet-based Subjective Logic to model uncertainty, and fuses semantic and visual evidence through a cautious conjunction rule to avoid overconfident fusion from correlated sources. Trained exclusively on one source domain and evaluated zero-shot across four unseen domains, SAEFS consistently outperforms state-of-the-art models both in prediction accuracy and reliability, improving the average C-index by 10.2\%.
Quantitative analyses further show that VQA-derived semantic features exhibit significantly lower cross-center divergence than pixel-derived features, highlighting their robustness for cross-center clinical applications. The code is available at \url{https://github.com/YuchengXing99/SAEFS}.

\end{abstract}

\section{Introduction}
\label{sec:intro}
Whole-slide images (WSIs) are gigapixel-scale digital scans of histopathology tissue sections that capture rich microscopic details, including cellular morphology, tissue architecture, and tumor micro-environments~\cite{kumar2020whole,song2023artificial}. These comprehensive pathological features enable clinicians to assess tumor progression and estimate patient prognosis, a task commonly referred to as survival analysis~\cite{mobadersany2018predicting,chen2022pan}. Since accurate survival prediction can provide critical guidance for personalized treatment planning and therapeutic decision-making, there has been growing interest in developing computational approaches to automate this process from WSIs~\cite{zhu2017wsisa,chen2022pan,jiang2025uncertainty}.

\begin{figure*}[t!]
  \centering
  \includegraphics[width=0.9\textwidth]{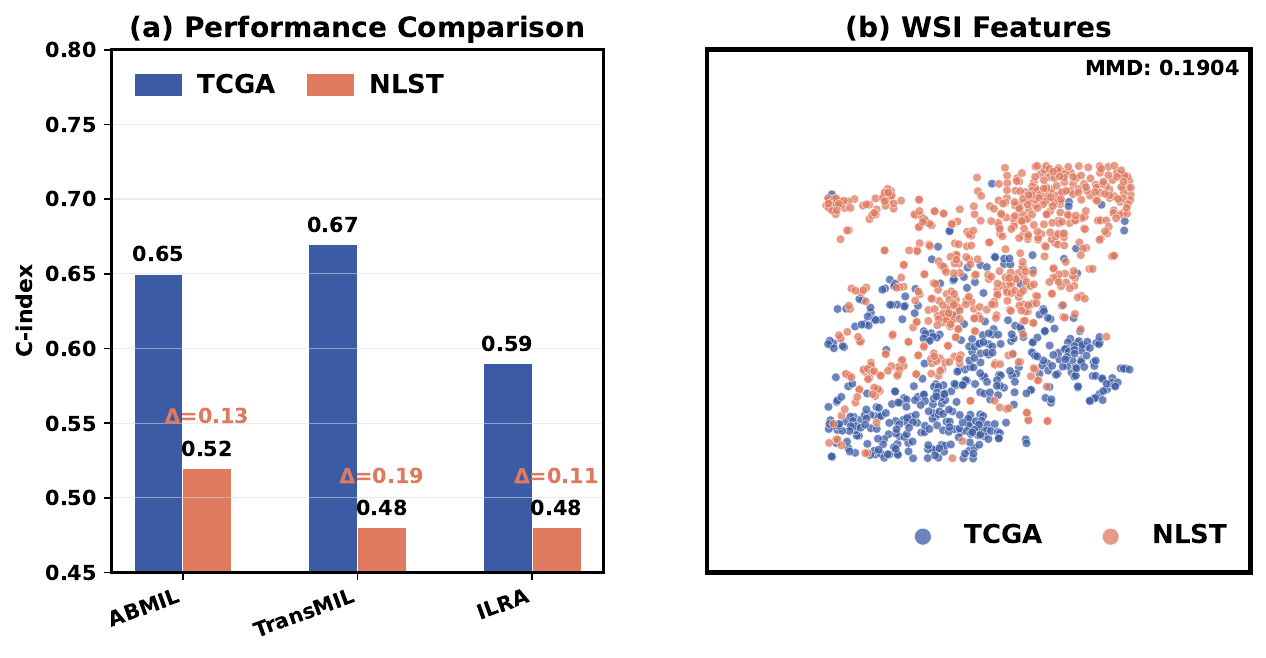}
  \caption{Domain-shift robustness analysis for lung adenocarcinoma under TCGA to NLST transfer. (a) C-index of ABMIL~\cite{ilse2018attention}, TransMIL~\cite{shao2021transmil}, and ILRA~\cite{xiang2023exploring} on the source domain (TCGA) and target domain (NLST), where $\Delta$ denotes the performance drop after transfer. (b) t-SNE visualization of cross-domain WSI features. Maximum Mean Discrepancy (MMD) values in (b) quantify the domain discrepancy in each feature space.}

  \label{fig:domain_shift}
\end{figure*}

To handle the extreme size of WSIs, attention-based multiple instance learning (MIL) has become the dominant paradigm~\cite{ilse2018attention,lu2021data,shao2021transmil}. By aggregating thousands of tissue patches into slide-level representations, these models capture complex morphological patterns, such as tumor cellularity and stromal architecture, that correlate with patient prognosis. However, these models are predominantly developed and validated within single-center cohorts, leaving their cross-center generalizability largely unexamined. In practice, WSIs collected from different centers inevitably exhibit substantial domain shifts due to heterogeneous staining protocols, diverse scanner hardware, and varying laboratory pipelines~\cite{tellez2019quantifying}. As illustrated in \cref{fig:domain_shift}(a), state-of-the-art MIL models such as TransMIL trained on The Cancer Genome Atlas (TCGA) suffer an average C-index drop of approximately 0.19 when evaluated zero-shot on the external cohort National Lung Screening Trial (NLST)~\cite{national2011reduced}, revealing a significant gap between benchmark and real-world performance.

Existing methods to address domain shift often rely on feature-level stain normalization~\cite{reinhard2002color,macenko2009method,vahadane2016structure}, which standardizes visual appearance through color space transformations but does not account for deeper structural and scanner-level variations~\cite{tellez2019quantifying}. Domain adaptation offers a more principled alternative. Supervised approaches explicitly align source and target distributions during training~\cite{liu2025hasd,shou2025graph}, while source-free methods~\cite{liang2020we,yang2021generalized} relax this constraint by adapting a pre-trained model using only unlabeled target data at inference time. Nevertheless, both paradigms fundamentally assume that a sufficient amount of target-domain data is accessible, either during training or adaptation. However, in real-world clinical deployment, the target data is typically unseen or completely unavailable due to privacy regulations and logistical constraints, making these approaches impractical.

We argue that this failure of cross-center transfer stems from a fundamental feature of entanglement: standard encoders conflate prognostically relevant morphology with domain-specific confounders such as staining intensity and scanner color profiles. Indeed, recent work has shown that current pathology foundation models encode center-of-origin signatures more strongly than biological signals~\cite{de2025current}, confirming that visual representations remain fragile to domain-level confounders (\cref{fig:domain_shift}(b)). In contrast, high-level pathology semantics, including descriptions of tumor grade, necrosis patterns, and the immune microenvironment, are inherently consistent across centers~\cite{saltz2018spatial}, mirroring clinical reality where a pathologist's expertise transfers because they reason via stable biological concepts rather than low-level pixel statistics. However, such semantic annotations have traditionally relied on expert pathologists to provide slide-level descriptions, making them prohibitively expensive and difficult to scale across large multi-center cohorts. These limitations motivate us to ask: \emph{can we build a survival model that automatically extracts domain-invariant semantic anchors and leverages them to alleviate cross-center distribution shift, without requiring any target-domain data?}

We answer this question affirmatively by proposing the \textbf{Semantic-Anchored Evidential Fusion Survival (SAEFS)} framework. To automatically obtain domain-invariant semantics at scale, SAEFS queries a pathology-specialized VQA model with template-structured prognostically relevant questions for each WSI, yielding a bounded, hallucination-resistant semantic space without manual annotation. Built upon these semantic anchors, SAEFS adopts a dual-stream architecture with a MIL stream for global visual context and a semantic-guided stream attending to prognosis-relevant patches, whose outputs are combined via Adaptive Evidence Mixing. The fused visual evidence and semantic evidence are then converted into belief masses and uncertainty through Dirichlet-based Subjective Logic~\cite{sensoy2018evidential}, and then integrated by a cautious conjunction rule~\cite{denoeux2008conjunctive} that accounts for evidence dependency. When evaluated zero-shot across four unseen centers, SAEFS consistently outperforms strong unimodal and multimodal baselines, with the most significant gains observed in high distribution-shift cohorts. Our contributions are:
\begin{enumerate}
    \item To the best of our knowledge, this is the first work to systematically study cross-center domain shift in the context of WSI-based survival analysis and propose a target-free solution that is robust to unseen target domain data.
    \item We introduce a template-based VQA pipeline that automatically extracts domain-invariant semantic features from a pathology VQA model, effectively anchoring prognostic feature extraction and offering a new paradigm for robust cross-center WSI analysis.
    \item We leverage cautious belief fusion to address the challenge of fusing correlated visual and semantic modalities, explicitly modeling inter-modal reliability and uncertainty to enable robust evidence integration under domain shift.
\end{enumerate}

\section{Related Work}
\label{sec:related_work}
\noindent\textbf{WSI-based Survival Prediction.}
Whole-slide survival research mainly builds on the multiple instance learning (MIL) paradigm with pure visual operation, in which a slide is represented as a set of patch-level instances and an aggregation module is used to produce a slide-level risk score~\cite{liu2026sparse,xing2026bridging}. ABMIL~\cite{ilse2018attention} introduced a gated attention mechanism for permutation-invariant aggregation, and CLAM~\cite{lu2021data} extended it with instance-level clustering and a data-efficient training pipeline. TransMIL~\cite{shao2021transmil} replaced the attention pooling with a Transformer encoder to capture inter-patch correlations, while DTFD-MIL~\cite{zhang2022dtfd} proposed a dual-tier feature distillation strategy for more discriminative bag representations. Beyond pure visual methods, multimodal approaches fuse histopathology features with genomic or transcriptomic data, e.g., MCAT~\cite{chen2021multimodal} employs cross-modal attention between pathology and omic embeddings, SurvPath~\cite{jaume2024modeling} models pathway-level interactions with a multimodal Transformer. More recently, the pathology foundation models pre-trained on large-scale slide collections, such as UNI~\cite{chen2024towards} and CONCH~\cite{lu2024avisionlanguage}, have provided stronger patch encoders. 
Despite these advances, most current methods are developed on single-center or pooled multi-center datasets and obscure the impact of domain distributional shift. Consequently, these methods suffer from substantial performance drops when applied across different data centers, as demonstrated in \cref{sec:experiments}.


\noindent\textbf{Domain Shift in Histopathology}
Domain shift in histopathology, driven by heterogeneous staining protocols and hardware-specific scanner characteristics, frequently degrades model generalization and introduces diagnostic bias. Traditional domain shift interventions primarily operate at the image level. Stain normalization techniques~\cite{macenko2009method,vahadane2016structure,reinhard2002color} attempt to map slides to a canonical color space, stain augmentation~\cite{tellez2019quantifying} seeks to desensitize models to color variations via synthetic perturbations. At the representation level, large-scale self-supervised learning (SSL) has emerged as a promising strategy for implicit domain generalization. State-of-the-art foundation models, such as CTransPath~\cite{wang2022transformer}, Phikon~\cite{filiot2023scaling}, and UNI~\cite{chen2024towards}, leverage massive multi-center datasets ($>$100k slides) to learn robust features. However, these encoders prioritize general feature quality over explicit domain invariance; as we demonstrate in \cref{sec:domain_analysis}, even these high-capacity feature spaces exhibit significant cross-center divergence, suggesting that scale alone does not guarantee domain robustness. General adaptation methods~\cite{wang2020tent,liang2025comprehensive,liu2025hasd,shou2025graph,liang2020we,yang2021generalized} aim to mitigate distribution shifts during deployment by dynamically updating model statistics. However, such approaches often require access to target-domain batches, a requirement frequently unmet in clinical workflows where slides are processed asynchronically and individually.

\noindent\textbf{Vision-Language Models and VQA in Pathology.}
The success of contrastive vision–language pre-training (CLIP~\cite{radford2021learning}) has catalyzed a wave of domain-specific adaptations in pathology. Early efforts focused on fine-tuning CLIP on pathology-specific pairs harvested from medical social media (PLIP~\cite{huang2023visual}) or large-scale curation of PubMed captions (CONCH~\cite{lu2024avisionlanguage}, QUILT-1M~\cite{ikezogwo2023quilt}). These semantic stability attributes of those models enable robust zero-shot classification and cross-modal retrieval by aligning visual features with textual descriptions. On the generative front, architectures like LLaVA-Med~\cite{li2023llava} and PathChat~\cite{lu2024multimodal} have extended these capabilities to biomedical VQA and open-ended slide-level dialogue. Furthermore, MI-Zero~\cite{lu2023visual} demonstrates the utility of CLIP-style embeddings for WSI-level classification using textual prompts. Despite these advancements, existing Vision–Language Models (VLMs) are primarily utilized for classification, retrieval, or descriptive report generation but lack application for cancer survival analysis.

\noindent\textbf{Uncertainty Estimation and Evidential Deep Learning.}
Quantifying predictive uncertainty is paramount for safe clinical deployment~\cite{huang2024evidential,huang2024review,huang2025evidential,xing2025dpsurv,xing2026evidentialfusionnetworkmultimodal}. While Bayesian approaches, such as MC Dropout~\cite{gal2016dropout} and Deep Ensembles~\cite{lakshminarayanan2017simple}, estimate uncertainty via stochastic sampling, they incur significant computational overhead during inference. In contrast, Evidential Deep Learning (EDL)~\cite{sensoy2018evidential} provides a deterministic, single-pass alternative by placing a Dirichlet prior over the categorical distribution. Grounded in Subjective Logic~\cite{jsang2018subjective}, EDL interprets the predicted concentration parameters as evidence supporting different outcomes, offering a principled calculus for reasoning under epistemic uncertainty. Recent efforts have extended EDL to multi-view classification~\cite{han2022trusted} and survival regression~\cite{jiang2025uncertainty}. However, conventional fusion operators such as Dempster's combination rule~\cite{dempster2008upper,shafer2020mathematical} assume source independence, which is frequently violated in domain shift where modalities share correlated information, causing overconfident predictions.

\section{Method}
\label{sec:method}
\begin{figure}[htp]
  \centering
  \includegraphics[width=\linewidth]{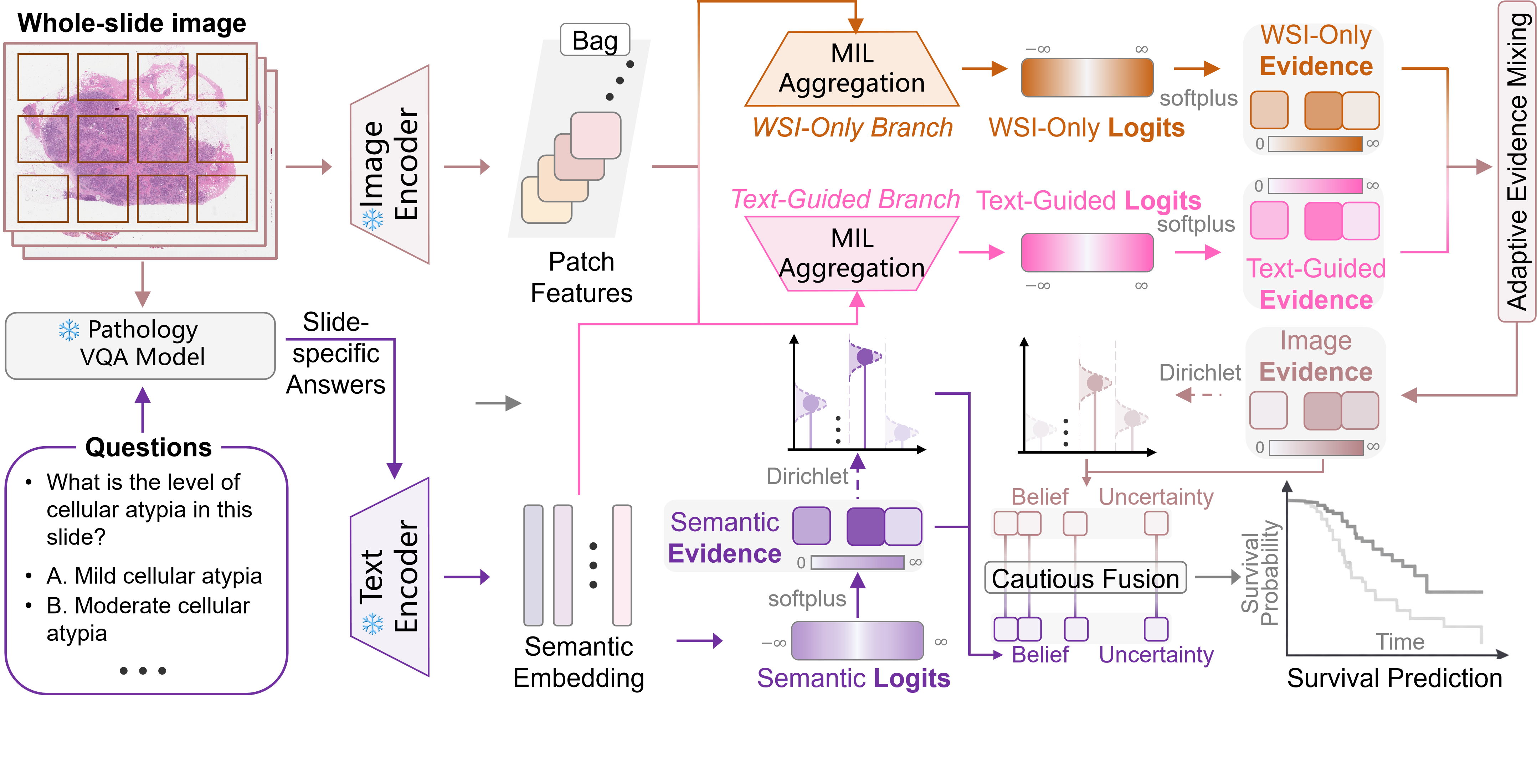}
  \caption{\textbf{Overview of the proposed framework.}
  Template-based VQA answers serve as semantic anchors to extract
  text-guided visual evidence from the WSI. A parallel WSI-only branch provides complementary visual evidence. The two visual evidence
  streams are adaptively mixed, then fused with pure text evidence
  through Dirichlet-based cautious belief fusion that preserves uncertainty under modality disagreement.}
  \label{fig:framework}
\end{figure}

The proposed SAEFS architecture is illustrated in Fig.~\ref{fig:framework}, composing semantic anchor generation, dual-stream semantic-anchored visual evidence extraction, semantic evidence extraction, and cautious belief fusion for survival predictions.
\subsection{Semantic Anchor Generation}
\label{sec:anchor_gen}
Semantic descriptions are more domain-robust in that they encode high-level pathology concepts that are more stable than raw visual features under distribution shift. We first exploit the VQA capability to automatically generate domain-robust semantic anchors.

\noindent \textbf{Template Question Generation.}
To define a structured semantic space, we construct a fixed set of $Q$ clinically grounded template questions in closed-form multiple-choice format, probing key pathology concepts including tumor morphology, grade, necrosis extent, stromal composition, and inflammatory infiltration. For example: \emph{``What is the level of cellular atypia in this slide? A. Mild cellular atypia. B. Moderate cellular atypia. C. Severe cellular atypia. D. Cellular atypia not assessable in this slide.''} These questions are generated using GPT-5.2 to capture clinically relevant prognostic factors in pathology. The closed-form design constrains the VQA model to select from predefined options rather than generating free-form text, yielding a bounded, hallucination-resistant semantic space with minimal linguistic variation. The complete set of 12 template questions, including all answer options and additional details, is provided in the Supplementary Material.

\noindent \textbf{Semantic Anchor Extraction via VQA.}
Given the predefined question set, we query a pathology-specialized VQA model for $\mathrm{WSI}_i$ to obtain the corresponding answers. 
The answers to all $Q$ questions are aggregated to form a structured slide-level semantic description. 
We denote the resulting semantic anchors for the $i$-th slide as
$\mathcal{T}_i = [\mathcal{T}_i^{1}, \dots, \mathcal{T}_i^{Q}]$.

\subsection{Dual-Stream WSI Evidence Extraction}
\label{sec:dual_stream}
To extract both prognostically relevant and complementary global survival information, we construct a text-guided branch that aggregates patch features under semantic guidance and a WSI-only branch that preserves holistic visual context, providing complementary global information.

\noindent Given the WSIs of a patient, we represent the processed patch-level features as a bag 
$\mathbf{X} = [\mathbf{x}_1, \dots, \mathbf{x}_N]^\top \in \mathbb{R}^{N \times D}$, 
where $\mathbf{x}_n \in \mathbb{R}^{D}$ denotes the feature vector of the $n$-th patch. These features are extracted using the image encoder of a vision-language model (VLM). 
In parallel, the semantic anchors for the $i$-th slide, $\mathcal{T}_i$, are encoded by the VLM's text encoder to obtain the semantic representations $\mathbf{P}_i = [\mathbf{p}_i^{1}, \dots, \mathbf{p}_i^{Q}]^\top \in \mathbb{R}^{Q \times D}$, where $\mathbf{p}_i^q \in \mathbb{R}^{D}$ denotes the feature vector of the $q$-th semantic anchor. Based on this instance representation, we derive two complementary branches:
\textbf{(a) Text-guided WSI Evidence branch:} selectively aggregates patch features under the guidance of semantic anchors, encouraging the model to focus on prognostically relevant regions and clinically meaningful patterns beyond raw visual statistics. 
\textbf{(b) WSI-only branch:} encodes holistic visual context without semantic conditioning, preserving global and information-rich representations across the entire slide.

\noindent\textbf{Text-Guided WSI Evidence Extraction.}
The embedding of each semantic anchor $\mathbf{p}^q$ serves as a query to retrieve prognostically relevant patches from the patch feature bag
$\mathbf{X}$ via similarity matching and weighted averaging:
\begin{equation}
  \mathbf{f}^q_{tg}
  = \sum_{i=1}^{N} \mathbf{x}_i \cdot
    \frac{
      \exp\bigl(\alpha \cdot \text{sim}(\mathbf{p}^q,\, \mathbf{x}_i)\bigr)
    }{
      \sum_{n=1}^{N} \exp\bigl(\alpha \cdot \text{sim}(\mathbf{p}^q,\, \mathbf{x}_n)\bigr)
    },
  \label{eq:coattn}
\end{equation}
where $\text{sim}(\cdot,\cdot)$ denotes cosine similarity and
$\alpha > 0$ is a fixed temperature hyperparameter. Applying Eq.~\eqref{eq:coattn} to all $Q$ semantic anchors yields a set of text-guided features $\{\mathbf{f}^q_{tg}\}_{q=1}^{Q}$. We aggregate the text-guided features via mean pooling and map the resulting representation to evidential outputs through a linear projection followed by a Softplus activation:
\begin{equation}
\mathbf{e}_{tg}
=
\mathrm{Softplus}\!\left(
\mathbf{W}_{tg}
\left(
\frac{1}{Q}
\sum_{q=1}^{Q}
\mathbf{f}^q_{tg}
\right)
+
\mathbf{b}_{tg}
\right) \in \mathbb{R}_{\ge 0}^{K},
\end{equation}
where $\mathbf{W}^{tg} \in \mathbb{R}^{K \times D}$ and 
$\mathbf{b}^{tg} \in \mathbb{R}^{K}$ are learnable parameters, and 
$\mathbf{e}^{tg}$ denotes a non-negative vector corresponding to $K$ discrete survival intervals.
Following the evidential learning framework, these non-negative outputs are interpreted as evidence that parameterizes the Dirichlet concentration parameters.


\noindent \textbf{WSI-only Evidence Extraction.}
To obtain a global representation of a WSI, we apply an attention-based pooling mechanism over patch features using
\begin{equation}\label{eq:attn_wo}
  a_i^{\text{wo}} =
    \frac{\exp\bigl((\mathbf{w}^{\text{wo}})^\top 
    \tanh(\mathbf{V}^{\text{wo}}\mathbf{x}_i)\bigr)}
         {\sum_{j=1}^{N}
          \exp\bigl((\mathbf{w}^{\text{wo}})^\top 
          \tanh(\mathbf{V}^{\text{wo}}\mathbf{x}_j)\bigr)},
  \qquad
  \mathbf{f}^{\text{wo}} = 
  \sum_{i=1}^{N} a_i^{\text{wo}} \, \mathbf{x}_i,
\end{equation}
where $\mathbf{V}^{\text{wo}} \in \mathbb{R}^{D \times D}$ and
$\mathbf{w}^{\text{wo}} \in \mathbb{R}^{D}$ are learnable parameters. The aggregated representation $\mathbf{f}^{\text{wo}}$ is then mapped to evidential outputs
via a linear projection followed by a Softplus activation:
\begin{equation}\label{eq:e_wo}
  \mathbf{e}^{\text{wo}}
  =
  \mathrm{Softplus}\!\left(
    \mathbf{W}^{\text{wo}} \mathbf{f}^{\text{wo}} 
    + \mathbf{b}^{\text{wo}}
  \right)
  \in \mathbb{R}_{\ge 0}^{K},
\end{equation}
where $\mathbf{W}^{\text{wo}} \in \mathbb{R}^{K \times D}$ 
and $\mathbf{b}^{\text{wo}} \in \mathbb{R}^{K}$ are learnable parameters.

\noindent \textbf{Adaptive Evidence Mixing:}
We combine the evidential outputs of the two branches using a mixing coefficient 
$\lambda \in [0,1]$:
\begin{equation}\label{eq:alpha_mix}
  \mathbf{e}_{\text{path}}
  =
  (1 - \lambda)\,\mathbf{e}^{\text{wo}}
  + \lambda\,\mathbf{e}^{\text{tg}},
\end{equation}
where $\lambda$ is a hyperparameter that controls the relative contribution of the WSI-only and text-guided branches, $\mathbf{e}_{\text{path}} \in \mathbb{R}_{\ge 0}^{K}$ 
denotes the mixed visual evidence.

\subsection{Text Evidence Extraction}
\label{sec:text}
Beyond guiding visual feature aggregation, the semantic representations themselves can serve as an additional modality for survival prediction. We thus map the semantic embeddings directly into evidential predictions to capture domain-stable prognostic signals.

\noindent To complement the visual evidence, we derive a purely semantic evidence stream from the anchor embeddings.
The $Q$ semantic anchor embeddings $\{\mathbf{p}_q\}_{q=1}^{Q}$
are aggregated via mean pooling and mapped to a non-negative evidence vector using Softplus activation:
\begin{equation}\label{eq:e_text}
  \mathbf{e}^{\text{text}}
  = \mathrm{Softplus}\!\left(
    \mathbf{W}^{\text{text}}
    \left(\frac{1}{Q}\sum_{q=1}^{Q} \mathbf{p}_q\right)
    + \mathbf{b}^{\text{text}}
  \right)
  \in \mathbb{R}_{\ge 0}^{K},
\end{equation}
where $\mathbf{W}^{\text{text}} \in \mathbb{R}^{K \times D}$ and
$\mathbf{b}^{\text{text}} \in \mathbb{R}^{K}$ are learnable parameters. 


\subsection{Cautious Belief Fusion}
\label{sec:fusion}
The textual and visual evidence streams are inherently dependent and combining them directly naively may amplify shared signals and lead to overconfident predictions. We therefore introduce cautious belief fusion, which preserves uncertainty and ensures more reliable predictions.

\noindent\textbf{Mapping evidence into belief.}
We now have two evidence sources:$\mathbf{e}_{\text{path}}$(\cref{eq:alpha_mix} from WSI source and $\mathbf{e}^{\text{text}}$ (\cref{eq:e_text}) from text source. Each evidence vector $\mathbf{e}$ defines a Dirichlet distribution
$\text{Dir}(\boldsymbol{\alpha})$ with parameters
$\boldsymbol{\alpha} = \mathbf{e} + \mathbf{1} \in \mathbb{R}_{\ge 0}^{K}$.
Following Subjective Logic~\cite{jsang2018subjective}, we convert the Dirichlet parameters into a \emph{belief--uncertainty pair}
$(\mathbf{b}, u)$:
\begin{equation}\label{eq:dir2bba}
  b_k = \frac{e_k}{S}, \qquad
  u   = \frac{K}{S},   \qquad
  S   = \sum_{k=1}^{K} \alpha_k,
\end{equation}
where $S$ is the Dirichlet strength and $K$ is the number of discrete survival intervals. The belief mass $b_k$ represents the evidence-supported probability for the $k$-th survival interval, while $u$ captures epistemic uncertainty due to lack of evidence.
By construction, $\sum_{k} b_k + u = 1$.
Applying \cref{eq:dir2bba} to $\mathbf{e}_{\text{path}}$ and
$\mathbf{e}^{\text{text}}$ yields two opinions:
$(\mathbf{b}_{\text{path}}, u_{\text{path}})$ and
$(\mathbf{b}_{\text{text}}, u_{\text{text}})$.

\noindent\textbf{Cautious conjunction rule.}
We fuse the two opinions using the \emph{cautious conjunction
rule}~\cite{denoeux2008conjunctive}, which operates through the
\emph{commonality function} $q_k = b_k + u$:
\begin{equation}\label{eq:cautious_q}
  q_k^{\text{fused}} = \min\bigl(
    q_k^{\text{path}},\; q_k^{\text{text}}
  \bigr),
  \qquad
  u^{\text{fused}} = \min\bigl(
    u_{\text{path}},\; u_{\text{text}}
  \bigr).
\end{equation}
The fused belief masses are recovered and normalized:
\begin{equation}\label{eq:cautious_b}
  \tilde{b}_k^{\text{fused}} = \max\bigl(
    q_k^{\text{fused}} - u^{\text{fused}},\; 0
  \bigr),
  \quad
  b_k^{\text{fused}} = \frac{\tilde{b}_k^{\text{fused}}}{Z},
  \quad
  u^{\text{fused}} \leftarrow \frac{u^{\text{fused}}}{Z},
\end{equation}
where $Z = \sum_{j} \tilde{b}_j^{\text{fused}} + u^{\text{fused}}$.
The key property of cautious fusion is uncertainty
preservation: if either modality assigns high uncertainty, the fused opinion retains that uncertainty rather than allowing the other modality's potentially unreliable confidence to dominate.

\subsection{Training Objective}
\label{sec:training}
\noindent Let $y_i \in \{0, \dots, K{-}1\}$ denote the discretized survival
interval and $c_i \in \{0, 1\}$ the censorship indicator
($c_i = 1$ if censored) for sample~$i$.
From the fused belief--uncertainty pair
$(\mathbf{b}^{\text{fused}}, u^{\text{fused}})$ obtained in
\cref{eq:cautious_b}, we recover the fused Dirichlet parameters:
\begin{equation}\label{eq:recover}
  S^{\text{fused}} = \frac{K}{u^{\text{fused}} + \epsilon},
  \qquad
  \alpha_k^{\text{fused}} = b_k^{\text{fused}} \cdot S^{\text{fused}} + 1,
\end{equation}
where $\epsilon$ is a small constant for numerical stability.
The expected categorical probability for the $k$-th interval is
$p_k = \alpha_k^{\text{fused}} / S^{\text{fused}}$.

\noindent\textbf{Survival loss.}
For uncensored samples ($c_i = 0$), we maximize the hazard
probability at the event interval; for censored samples ($c_i = 1$), we maximize the survival probability beyond the last observed interval:
\begin{equation}\label{eq:loss_surv}
\mathcal{L}_{\text{surv}} =
-\sum_{i=1}^{N}
\Bigl[
(1-c_i)\log p_{y_i}^{(i)}
+
c_i \log \sum_{k>y_i} p_k^{(i)}
\Bigr].
\end{equation}
The first term encourages the model to assign a high probability to the interval where the event occurs. The second term ensures that for censored patients, who are known to survive beyond interval $y_i$, the model assigns a high cumulative probability to the subsequent intervals.

\noindent\textbf{KL regularizer.}
To prevent the model from accumulating spurious evidence for
incorrect intervals, we add a KL-divergence regularizer that
penalizes deviation from a non-informative uniform
Dirichlet~\cite{sensoy2018evidential}:
\begin{equation}\label{eq:loss_kl}
\mathcal{L}_{\text{KL}} =
\sum_{i=1}^{N}
\mathrm{KL}\bigl[
\mathrm{Dir}(\tilde{\boldsymbol{\alpha}}^{(i)})
\;\|\;
\mathrm{Dir}(\mathbf{1})
\bigr],
\end{equation}
where $\tilde{\alpha}_k^{(i)} = \alpha_k^{(i)} - e_k^{(i)}
\cdot \mathbf{1}[k = y_i] + 1$ removes the evidence for the
ground-truth interval before penalization, so that only the
evidence assigned to \emph{incorrect} intervals is regularized
toward zero. The total loss is:
\begin{equation}\label{eq:loss_total}
  \mathcal{L} = \mathcal{L}_{\text{surv}}
  + \lambda_t \cdot \mathcal{L}_{\text{KL}},
\end{equation}
where $\lambda_t = \min(1,\; t / T_{\text{anneal}})$ is an annealing coefficient that linearly increases from $0$ to $1$ over the first $T_{\text{anneal}}$ training steps, allowing the model to freely accumulate evidence in early training before the regularizer takes full effect.

\section{Experiments}
\label{sec:experiments}

\begin{table}[t!]
\centering
\caption{Cross-center survival prediction results (TCGA $\rightarrow$ CPTAC/NLST) under unimodal settings.
$C$: C-index$\uparrow$, $S$: IBS$\downarrow$, $I$: INBLL$\downarrow$.
Results are reported as mean $\pm$ std over three runs.
Best results are in \textbf{bold} and our results are \colorbox{blue!5}{color-coded}.}
\label{tab:cross_domain}
\renewcommand{\arraystretch}{1.1}
\resizebox{\textwidth}{!}{%
\setlength{\tabcolsep}{5pt}
\begin{tabular}{l | ccc | ccc | ccc | ccc | ccc}
\toprule
\multirow{2}{*}{\textbf{Method}} 
  & \multicolumn{3}{c|}{\textbf{CPTAC-LUAD}} 
  & \multicolumn{3}{c|}{\textbf{CPTAC-UCEC}} 
  & \multicolumn{3}{c|}{\textbf{CPTAC-KIRC}} 
  & \multicolumn{3}{c|}{\textbf{NLST-LUAD}}
  & \multicolumn{3}{c}{\textbf{Avg}} \\
\cmidrule(lr){2-4} \cmidrule(lr){5-7} 
\cmidrule(lr){8-10} \cmidrule(lr){11-13}
\cmidrule(lr){14-16}
& $C\uparrow$ & $S\downarrow$ & $I\downarrow$
& $C\uparrow$ & $S\downarrow$ & $I\downarrow$
& $C\uparrow$ & $S\downarrow$ & $I\downarrow$
& $C\uparrow$ & $S\downarrow$ & $I\downarrow$
& $C\uparrow$ & $S\downarrow$ & $I\downarrow$ \\
\midrule
ABMIL & 0.464{\scriptsize$\pm$0.01} & 0.702{\scriptsize$\pm$0.01} & 1.962{\scriptsize$\pm$0.06} & 0.534{\scriptsize$\pm$0.01} & 0.685{\scriptsize$\pm$0.00} & 2.984{\scriptsize$\pm$0.03} & 0.600{\scriptsize$\pm$0.01} & 0.814{\scriptsize$\pm$0.01} & 3.412{\scriptsize$\pm$0.20} & 0.535{\scriptsize$\pm$0.01} & 0.706{\scriptsize$\pm$0.00} & 2.112{\scriptsize$\pm$0.01} & 0.533 & 0.727 & 2.617 \\
TransMIL & 0.500{\scriptsize$\pm$0.02} & 0.946{\scriptsize$\pm$0.00} & 5.081{\scriptsize$\pm$0.05} & 0.636{\scriptsize$\pm$0.04} & 0.833{\scriptsize$\pm$0.03} & 5.307{\scriptsize$\pm$0.23} & 0.622{\scriptsize$\pm$0.01} & 0.868{\scriptsize$\pm$0.00} & 5.431{\scriptsize$\pm$0.15} & 0.516{\scriptsize$\pm$0.03} & 0.908{\scriptsize$\pm$0.00} & 4.728{\scriptsize$\pm$0.02} & 0.568 & 0.889 & 5.136 \\
DSMIL & 0.469{\scriptsize$\pm$0.00} & 0.613{\scriptsize$\pm$0.00} & 1.575{\scriptsize$\pm$0.02} & 0.607{\scriptsize$\pm$0.02} & 0.662{\scriptsize$\pm$0.02} & 2.199{\scriptsize$\pm$0.11} & 0.637{\scriptsize$\pm$0.00} & 0.772{\scriptsize$\pm$0.00} & 2.596{\scriptsize$\pm$0.07} & 0.534{\scriptsize$\pm$0.02} & 0.525{\scriptsize$\pm$0.02} & 1.308{\scriptsize$\pm$0.04} & 0.562 & 0.643 & 1.920 \\
ILRA & 0.519{\scriptsize$\pm$0.05} & 0.974{\scriptsize$\pm$0.03} & 6.553{\scriptsize$\pm$0.13} & 0.638{\scriptsize$\pm$0.06} & 0.877{\scriptsize$\pm$0.00} & 6.899{\scriptsize$\pm$0.45} & 0.616{\scriptsize$\pm$0.01} & 0.904{\scriptsize$\pm$0.00} & 7.071{\scriptsize$\pm$0.01} & 0.596{\scriptsize$\pm$0.02} & 0.865{\scriptsize$\pm$0.02} & 5.745{\scriptsize$\pm$0.07} & 0.593 & 0.905 & 6.567 \\
BayesMIL & 0.402{\scriptsize$\pm$0.00} & 0.619{\scriptsize$\pm$0.02} & 1.578{\scriptsize$\pm$0.05} & \textbf{0.702{\scriptsize$\pm$0.02}} & 0.658{\scriptsize$\pm$0.00} & 2.322{\scriptsize$\pm$0.00} & 0.564{\scriptsize$\pm$0.00} & 0.787{\scriptsize$\pm$0.00} & 2.816{\scriptsize$\pm$0.05} & 0.503{\scriptsize$\pm$0.00} & 0.581{\scriptsize$\pm$0.01} & 1.484{\scriptsize$\pm$0.02} & 0.543 & 0.661 & 2.050 \\
UMSA & 0.435{\scriptsize$\pm$0.03} & 0.709{\scriptsize$\pm$0.02} & 1.979{\scriptsize$\pm$0.09} & 0.540{\scriptsize$\pm$0.03} & 0.700{\scriptsize$\pm$0.05} & 3.005{\scriptsize$\pm$0.20} & 0.605{\scriptsize$\pm$0.01} & 0.816{\scriptsize$\pm$0.01} & 3.390{\scriptsize$\pm$0.13} & 0.530{\scriptsize$\pm$0.02} & 0.688{\scriptsize$\pm$0.03} & 2.020{\scriptsize$\pm$0.15} & 0.528 & 0.728 & 2.599 \\
ACMIL & 0.405{\scriptsize$\pm$0.10} & 0.679{\scriptsize$\pm$0.00} & 1.895{\scriptsize$\pm$0.02} & 0.644{\scriptsize$\pm$0.00} & 0.721{\scriptsize$\pm$0.00} & 2.723{\scriptsize$\pm$0.06} & 0.642{\scriptsize$\pm$0.00} & 0.790{\scriptsize$\pm$0.00} & 2.791{\scriptsize$\pm$0.01} & 0.559{\scriptsize$\pm$0.00} & 0.793{\scriptsize$\pm$0.02} & 2.577{\scriptsize$\pm$0.10} & 0.563 & 0.746 & 2.497 \\
OTSurv & 0.529{\scriptsize$\pm$0.06} & 0.570{\scriptsize$\pm$0.12} & 1.432{\scriptsize$\pm$0.32} & 0.612{\scriptsize$\pm$0.00} & 0.702{\scriptsize$\pm$0.00} & 2.035{\scriptsize$\pm$0.04} & 0.652{\scriptsize$\pm$0.01} & 0.724{\scriptsize$\pm$0.03} & 2.110{\scriptsize$\pm$0.07} & 0.644{\scriptsize$\pm$0.01} & 0.523{\scriptsize$\pm$0.00} & 1.307{\scriptsize$\pm$0.01} & 0.609 & 0.630 & 1.721 \\
\midrule
\rowcolor{blue!5}
Ours 
& \textbf{0.663{\scriptsize$\pm$0.02}} & \textbf{0.376{\scriptsize$\pm$0.01}} & \textbf{0.910{\scriptsize$\pm$0.03}} 
& 0.682{\scriptsize$\pm$0.02} & \textbf{0.270{\scriptsize$\pm$0.00}} & \textbf{0.703{\scriptsize$\pm$0.00}} 
& \textbf{0.677{\scriptsize$\pm$0.03}} & \textbf{0.272{\scriptsize$\pm$0.01}} & \textbf{0.700{\scriptsize$\pm$0.02}} 
& \textbf{0.662{\scriptsize$\pm$0.02}} & \textbf{0.278{\scriptsize$\pm$0.01}} & \textbf{0.709{\scriptsize$\pm$0.02}}
& \textbf{0.671} & \textbf{0.299} & \textbf{0.755} \\
\bottomrule
\end{tabular}
}
\end{table}

\subsection{Experimental Setup}
\label{sec:setup}

\noindent\textbf{Datasets.}
We use The Cancer Genome Atlas (TCGA) database ~\cite{tomczak2015review} as the source domain for training and test the trained model on four external cohorts from independent centers to evaluate cross-domain generalization under the zero-shot setting. TCGA covers three cancer types: 
Lung Adenocarcinoma (LUAD), Uterine Corpus Endometrial Carcinoma (UCEC), and Kidney Renal Clear Cell Carcinoma (KIRC).
The four external cohorts include CPTAC~\cite{edwards2015cptac} (CPTAC-LUAD, CPTAC-UCEC, and CPTAC-KIRC), and NLST~\cite{national2011reduced} (NLST-LUAD). These cohorts exhibit substantial distribution shifts in staining protocols, scanner hardware, and tissue preparation pipelines, posing a challenging benchmark for domain robustness. No data from the target cohorts are used during training.

\noindent\textbf{Evaluation Metrics.}
Following \cite{huang2025esurvfusion}, we adopt three complementary metrics for survival prediction assessment: concordance index (C-index) measures the discriminative ability of the model by computing the fraction of concordant pairs among all comparable patient pairs; Integrated Brier Score (IBS) evaluates the calibration and sharpness of the predicted survival probabilities over time; and Integrated Negative Binomial Log-Likelihood (INBLL) assesses the overall probabilistic fit of the predicted survival distributions.

\noindent\textbf{Baselines.}
We compare SAEFS against two groups of state-of-the-art methods.
\textbf{(1) Unimodal methods:}
ABMIL~\cite{ilse2018attention},
TransMIL~\cite{shao2021transmil},
DSMIL~\cite{li2021dual},
ILRA~\cite{xiang2023exploring},
BayesMIL~\cite{cui2023bayes},
UMSA~\cite{jiang2025uncertainty},
ACMIL~\cite{zhang2024attention}, and
OTSurv~\cite{ren2025otsurv}.
\textbf{(2) Multimodal methods} that fuse WSI features with VQA-derived textual representations:
MCAT~\cite{chen2021multimodal},
MOTCAT~\cite{xu2023multimodal},
SurvPath~\cite{jaume2024modeling}, and
PS3~\cite{raza2025ps3}.
All baselines are trained on the same TCGA splits with the CONCH~\cite{lu2024avisionlanguage} patch encoder for fair comparison and evaluated zero-shot on external cohorts. Methods that use target-domain data for training, e.g., domain adaptation, are excluded for fairness. See the Supplementary Material for implementation details.

\subsection{Comparison with State-of-the-Art Methods}
\label{sec:sota}

\noindent\textbf{Comparison with Unimodal Methods.}
Table~\ref{tab:cross_domain} summarizes the zero-shot cross-center performance of our method against eight state-of-the-art WSI-only survival prediction methods. Our method achieves the highest average C-index of 0.671, exceeding the strongest baseline by +0.062. The improvement is particularly notable on the high-shift CPTAC-LUAD cohort, where SAEFS attains a C-index of 0.663 compared to 0.529 for the best competing method. 
\textbf{Notably, the performance gains become more pronounced as the domain shift increases, indicating that SAEFS is particularly effective under severe cross-center distribution shifts.}
In terms of prediction uncertainty, our method also achieves the lowest average IBS (0.299) and INBLL (0.755), indicating more accurate and reliable survival estimation under domain shift. 
These results suggest that combining semantic anchoring with evidential fusion improves cross-domain robustness compared with purely visual approaches.

\begin{table}[t]
\centering
\caption{Cross-domain survival prediction results for multimodal methods (TCGA $\rightarrow$ CPTAC/NLST). 
$C$: C-index$\uparrow$, $S$: IBS$\downarrow$, $I$: INBLL$\downarrow$. Results are reported as mean $\pm$ std over three runs.
Best results are in \textbf{bold} and our results are \colorbox{blue!5}{color-coded}.}
\label{tab:cross_domain_multimodal}
\renewcommand{\arraystretch}{1.1}
\resizebox{\textwidth}{!}{%
\setlength{\tabcolsep}{3pt}
\begin{tabular}{l | ccc | ccc | ccc | ccc | ccc}
\toprule
\multirow{2}{*}{\textbf{Method}} 
  & \multicolumn{3}{c|}{\textbf{CPTAC-LUAD}} 
  & \multicolumn{3}{c|}{\textbf{CPTAC-UCEC}} 
  & \multicolumn{3}{c|}{\textbf{CPTAC-KIRC}} 
  & \multicolumn{3}{c|}{\textbf{NLST-LUAD}}
  & \multicolumn{3}{c}{\textbf{Avg}} \\
\cmidrule(lr){2-4} \cmidrule(lr){5-7} 
\cmidrule(lr){8-10} \cmidrule(lr){11-13}
\cmidrule(lr){14-16}
& $C\uparrow$ & $S\downarrow$ & $I\downarrow$
& $C\uparrow$ & $S\downarrow$ & $I\downarrow$
& $C\uparrow$ & $S\downarrow$ & $I\downarrow$
& $C\uparrow$ & $S\downarrow$ & $I\downarrow$
& $C\uparrow$ & $S\downarrow$ & $I\downarrow$ \\
\midrule
MCAT      
& 0.440{\scriptsize$\pm$0.05} & 0.286{\scriptsize$\pm$0.00} & 0.873{\scriptsize$\pm$0.03} 
& 0.602{\scriptsize$\pm$0.07} & \textbf{0.214}{\scriptsize$\pm$0.01} & 0.824{\scriptsize$\pm$0.05} 
& 0.591{\scriptsize$\pm$0.03} & 0.217{\scriptsize$\pm$0.01} & 0.793{\scriptsize$\pm$0.03} 
& 0.460{\scriptsize$\pm$0.05} & 0.266{\scriptsize$\pm$0.04} & 0.822{\scriptsize$\pm$0.11}
& 0.523 & 0.246 & 0.828 \\
MOTCAT    
& 0.436{\scriptsize$\pm$0.03} & \textbf{0.274}{\scriptsize$\pm$0.02} & \textbf{0.844}{\scriptsize$\pm$0.09} 
& 0.635{\scriptsize$\pm$0.06} & 0.217{\scriptsize$\pm$0.00} & 0.791{\scriptsize$\pm$0.06} 
& 0.555{\scriptsize$\pm$0.04} & \textbf{0.211}{\scriptsize$\pm$0.01} & 0.733{\scriptsize$\pm$0.04} 
& 0.510{\scriptsize$\pm$0.01} & \textbf{0.245}{\scriptsize$\pm$0.02} & \textbf{0.708}{\scriptsize$\pm$0.05}
& 0.534 & \textbf{0.237} & 0.769 \\
SurvPath  
& 0.426{\scriptsize$\pm$0.07} & 0.289{\scriptsize$\pm$0.01} & 0.894{\scriptsize$\pm$0.04} 
& 0.572{\scriptsize$\pm$0.02} & 0.220{\scriptsize$\pm$0.01} & 0.842{\scriptsize$\pm$0.04} 
& 0.627{\scriptsize$\pm$0.04} & 0.213{\scriptsize$\pm$0.01} & 0.756{\scriptsize$\pm$0.04} 
& 0.431{\scriptsize$\pm$0.01} & 0.279{\scriptsize$\pm$0.02} & 0.850{\scriptsize$\pm$0.07}
& 0.514 & 0.250 & 0.836 \\
PS3       
& 0.455{\scriptsize$\pm$0.04} & 0.287{\scriptsize$\pm$0.01} & 1.023{\scriptsize$\pm$0.06} 
& 0.666{\scriptsize$\pm$0.02} & 0.218{\scriptsize$\pm$0.02} & 1.058{\scriptsize$\pm$0.09} 
& 0.596{\scriptsize$\pm$0.04} & 0.229{\scriptsize$\pm$0.01} & 1.020{\scriptsize$\pm$0.06} 
& 0.529{\scriptsize$\pm$0.04} & 0.254{\scriptsize$\pm$0.01} & 0.872{\scriptsize$\pm$0.04}
& 0.562 & 0.247 & 0.993 \\
\midrule
\rowcolor{blue!5}
Ours 
& \textbf{0.663}{\scriptsize$\pm$0.02} & 0.376{\scriptsize$\pm$0.01} & 0.910{\scriptsize$\pm$0.03} 
& \textbf{0.682}{\scriptsize$\pm$0.02} & 0.270{\scriptsize$\pm$0.00} & \textbf{0.703}{\scriptsize$\pm$0.00} 
& \textbf{0.677}{\scriptsize$\pm$0.03} & 0.272{\scriptsize$\pm$0.01} & \textbf{0.700}{\scriptsize$\pm$0.02} 
& \textbf{0.662}{\scriptsize$\pm$0.02} & 0.278{\scriptsize$\pm$0.01} & 0.709{\scriptsize$\pm$0.02}
& \textbf{0.671} & 0.299 & \textbf{0.755} \\
\bottomrule
\end{tabular}
}
\end{table}

\noindent\textbf{Comparison with Multimodal Methods.}
Table~\ref{tab:cross_domain_multimodal} compares SAEFS with several multimodal survival prediction methods that combine WSI features with semantic representations. 
SAEFS achieves the highest average C-index of 0.671, exceeding the strongest competing method by +0.109. 
Notably, while multimodal baselines yield slightly lower IBS scores, we attribute this to inter-modal correlation: naive concatenation-based fusion implicitly double-counts shared prognostic signals between visual and semantic modalities, producing overconfident predictions that appear sharp but are poorly calibrated. In contrast, SAEFS explicitly models per-modal reliability through cautious belief fusion, avoiding such overconfidence and achieving the best average INBLL of 0.755 alongside substantially stronger discriminative power. The calibration analysis in Sec.~\ref{sec:risk_calibration} corroborates this interpretation, showing that SAEFS produces well-calibrated survival estimates while baseline models exhibit systematic deviation between predicted and observed probabilities.

\noindent\textbf{Model complexity.}
Although SAEFS adds a semantic branch, its trainable fusion module remains lightweight, containing only 270K parameters, compared with 659K for ABMIL, 2.41M for TransMIL, and 2.74M for MCAT. Semantic anchors are generated once per slide using the VQA model and cached offline, introducing no additional computational cost during fusion-model training or inference.

\begin{table}[t!]
\centering
\caption{\textbf{Ablation study (average over four external datasets).}
TCGA$\rightarrow$CPTAC/NLST zero-shot evaluation.
A1 removes the text-guided visual stream (WSI-only, $\lambda=0$).
A2 removes the global WSI branch and uses only text-guided visual features (TG-WSI only, $\lambda=1$).
A3 uses only semantic features (Text-only).
A4 replaces evidential fusion with simple feature concatenation.
A5 replaces the cautious fusion rule with the Dempster--Shafer rule.
$\Delta$ is computed w.r.t. the full model.}
\label{tab:ablation_avg}

\renewcommand{\arraystretch}{1.15}
\setlength{\tabcolsep}{4pt}
\resizebox{0.8\columnwidth}{!}{
\begin{tabular}{c l cc cc cc}
\toprule
& &
\multicolumn{2}{c}{\textbf{C-index} $\uparrow$} &
\multicolumn{2}{c}{\textbf{IBS} $\downarrow$} &
\multicolumn{2}{c}{\textbf{INBLL} $\downarrow$} \\
\cmidrule(lr){3-4}
\cmidrule(lr){5-6}
\cmidrule(lr){7-8}
Action & Configuration
& Value & $\Delta$
& Value & $\Delta$
& Value & $\Delta$ \\
\midrule

A1 & WSI-only ($\lambda=0$)
& 0.5980 & -0.0860
& 0.2933 & -0.0098
& 0.7452 & -0.0196 \\

A2 & TG-WSI only ($\lambda=1$)
& 0.6408 & -0.0432
& 0.2964 & -0.0067
& 0.7490 & -0.0158 \\

A3 & Text-only
& 0.5327 & -0.1513
& 0.3769 & +0.0738
& 0.9417 & +0.1769 \\

A4 & Concatenation fusion
& 0.5310 & -0.1530
& 0.2119 & -0.0912
& 0.7551 & -0.0098 \\

A5 & Dempster fusion
& 0.5758 & -0.1082
& 0.3113 & +0.0082
& 0.7824 & +0.0176 \\
\rowcolor{blue!5}
& SAEFS (Full model)
& 0.6840 & ---
& 0.3031 & ---
& 0.7648 & --- \\
\bottomrule
\end{tabular}
}
\end{table}
\subsection{Ablation Study}
\label{sec:ablation}

To assess the effectiveness of each component of SAEFS, we conduct a systematic ablation study averaged over the four external datasets (Table~\ref{tab:ablation_avg}). 
The ablation variants progressively modify the visual streams and the fusion mechanism to analyze their individual impact on cross-center survival prediction.

\noindent\textbf{Effect of visual and semantic modalities.}
We first examine the role of the dual visual branches. 
Removing the text-guided visual stream (A1: WSI-only) leads to a notable C-index drop of $-0.0860$, indicating that semantic guidance provides important cues for identifying prognostically relevant regions under domain shift. 
Using only the text-guided visual stream without the global WSI branch (A2: TG-WSI only) also degrades performance ($-0.0432$), suggesting that global visual context remains complementary to the semantically guided features. 
Finally, the text-only variant (A3) yields the largest degradation ($-0.1513$), showing that VQA-derived semantics alone, while relatively domain-stable, are insufficient for fine-grained survival discrimination without visual grounding.

\noindent\textbf{Effect of cautious belief fusion.}
We next analyze the fusion mechanism for visual and semantic evidence. 
Replacing the evidential fusion with a simple concatenation-based predictor (A4) significantly degrades performance ($-0.1530$), highlighting the importance of uncertainty-aware fusion when combining heterogeneous modalities under domain shift. 
Furthermore, substituting the cautious conjunction rule with the conventional Dempster's combination rule (A5) also reduces performance ($-0.1082$) and slightly worsens IBS and INBLL. 
This suggests that the independence assumption underlying Dempster's combination rule may be violated in our setting, where visual and semantic evidence can be strongly correlated. 
The cautious fusion rule, which is designed to handle dependent evidence more conservatively, therefore yields more reliable predictions. Overall, the full SAEFS model achieves the best balance between discrimination and calibration, confirming the importance of both the dual-stream semantic guidance and the cautious evidential fusion design.

\noindent\textbf{Attribution of performance gains.}
We conduct four control experiments to examine whether the improvement arises from the proposed fusion strategy or simply from using a strong VQA model (Table~\ref{tab:attribution}).
Directly concatenating the same VQA features with OTSurv does not improve over unimodal OTSurv, suggesting that the additional VQA information alone is insufficient.
Replacing the semantic anchors with random text leads to a substantial performance drop, highlighting the importance of meaningful semantic guidance.
Moreover, replacing ALPaCA with SlideChat yields comparable C-index performance, showing that SAEFS is not restricted to a particular VQA backbone.
Combined with the text-only ablation (A3), these results support that the cross-center improvement mainly comes from the uncertainty-aware dual-stream fusion rather than the choice of the VQA model.

\begin{table}[t]
\centering
\caption{Control experiments on CPTAC-UCEC under zero-shot transfer
(TCGA$\rightarrow$CPTAC-UCEC). C: C-index$\uparrow$;
S: IBS$\downarrow$; I: INBLL$\downarrow$.
The best result for each metric is shown in bold.}
\label{tab:attribution}
\begin{tabular}{lccc}
\toprule
Method & C $\uparrow$ & S $\downarrow$ & I $\downarrow$ \\
\midrule
OTSurv (VQA concat)                  & 0.596 & 0.600 & 1.873 \\
SAEFS (random anchors)               & 0.504 & 0.532 & 1.424 \\
SAEFS (SlideChat \cite{Chen_2025_CVPR}) & 0.678 & 0.535 & 1.432 \\
\textbf{SAEFS (Ours)}                & \textbf{0.682} & \textbf{0.270} & \textbf{0.703} \\
\bottomrule
\end{tabular}
\end{table}

\noindent\textbf{Sensitivity to the number of questions.}
We further vary the number of template questions and observe a consistent improvement as coverage increases ($Q{=}4$: 0.633, $Q{=}8$: 0.660, $Q{=}12$: 0.682), indicating that a broader set of prognostic factors provides more complete semantic coverage.


\begin{figure*}[t!]
  \centering
  \includegraphics[width=\textwidth]{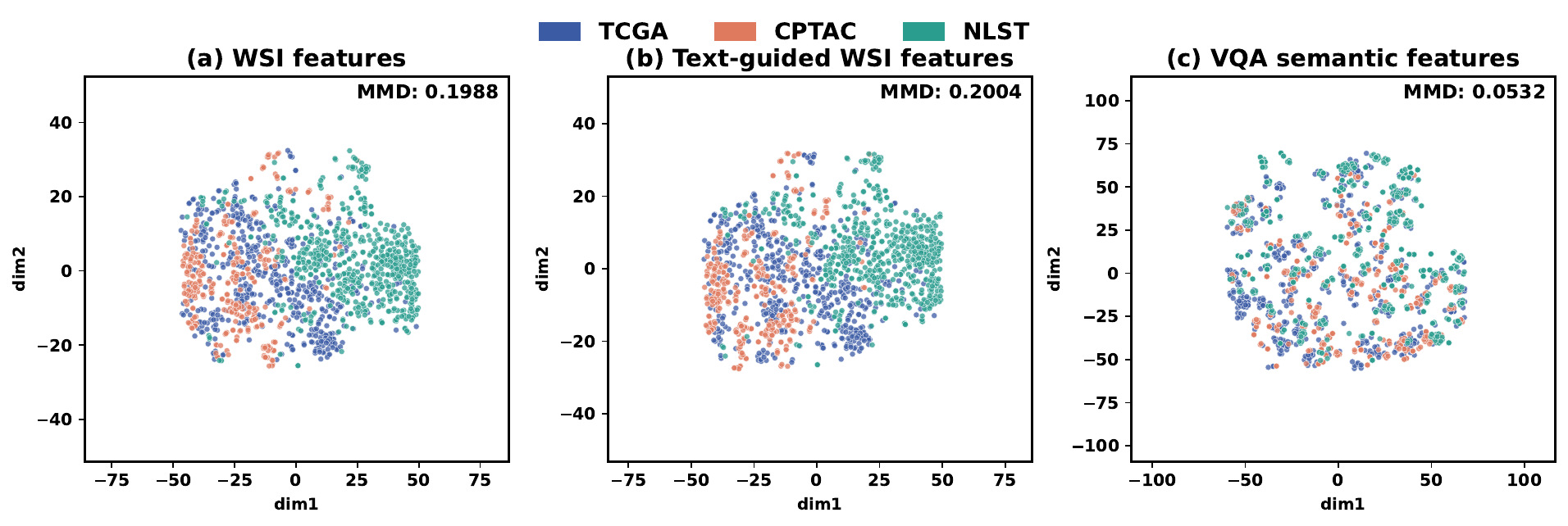}
  \caption{\textbf{Cross-center feature distributions under domain shift.}
            t-SNE visualization of three feature representations extracted from TCGA, CPTAC, and NLST: (a) raw WSI features, (b) text-guided WSI features, and (c) VQA-derived semantic features. 
            Colors indicate different centers. Semantic features exhibit substantially better cross-center alignment with a much lower Maximum Mean Discrepancy (MMD).}
  \label{fig:three_distributions}
\end{figure*}
\subsection{Analysis of Domain Robustness}
\label{sec:domain_analysis}

\noindent\textbf{Cross-Center Feature Divergence.}
To empirically validate the effectiveness of semantic information for cross-center generalization, we quantify the distributional divergence between the source domain (TCGA) and external cohorts (CPTAC and NLST) using Maximum Mean Discrepancy (MMD)~\cite{gretton2012kernel}. 
We compare three feature representations: (a) raw WSI features extracted from the patch encoder, (b) text-guided WSI features produced by the semantic cross-attention (Eq.~\ref{eq:coattn}), and (c) VQA-derived semantic features from the text encoder.

As illustrated in Fig.~\ref{fig:three_distributions}, the raw WSI features (Fig.~\ref{fig:three_distributions}a) exhibit clear distributional separation between TCGA and the external cohorts, with an MMD of 0.1988. 
The text-guided WSI features (Fig.~\ref{fig:three_distributions}b) show a comparable level of divergence (MMD = 0.2004), indicating that although semantic guidance alters the visual representation, the underlying domain shift in pixel-level features remains substantial. 
In contrast, the VQA semantic features (Fig.~\ref{fig:three_distributions}c) display strong overlap across centers, yielding a dramatically lower MMD of 0.0532—approximately a 73\% reduction compared with the raw visual features. 
This result suggests that high-level pathology semantics features derived from template-structured VQA are substantially more domain-invariant than pixel-derived representations, validating the improved cross-center robustness of SAEFS.

\subsection{Risk Stratification and Calibration Analysis}
\label{sec:risk_calibration}

\begin{figure*}[t!]
\centering

\begin{subfigure}{0.24\textwidth}
    \centering
    \includegraphics[width=\linewidth]{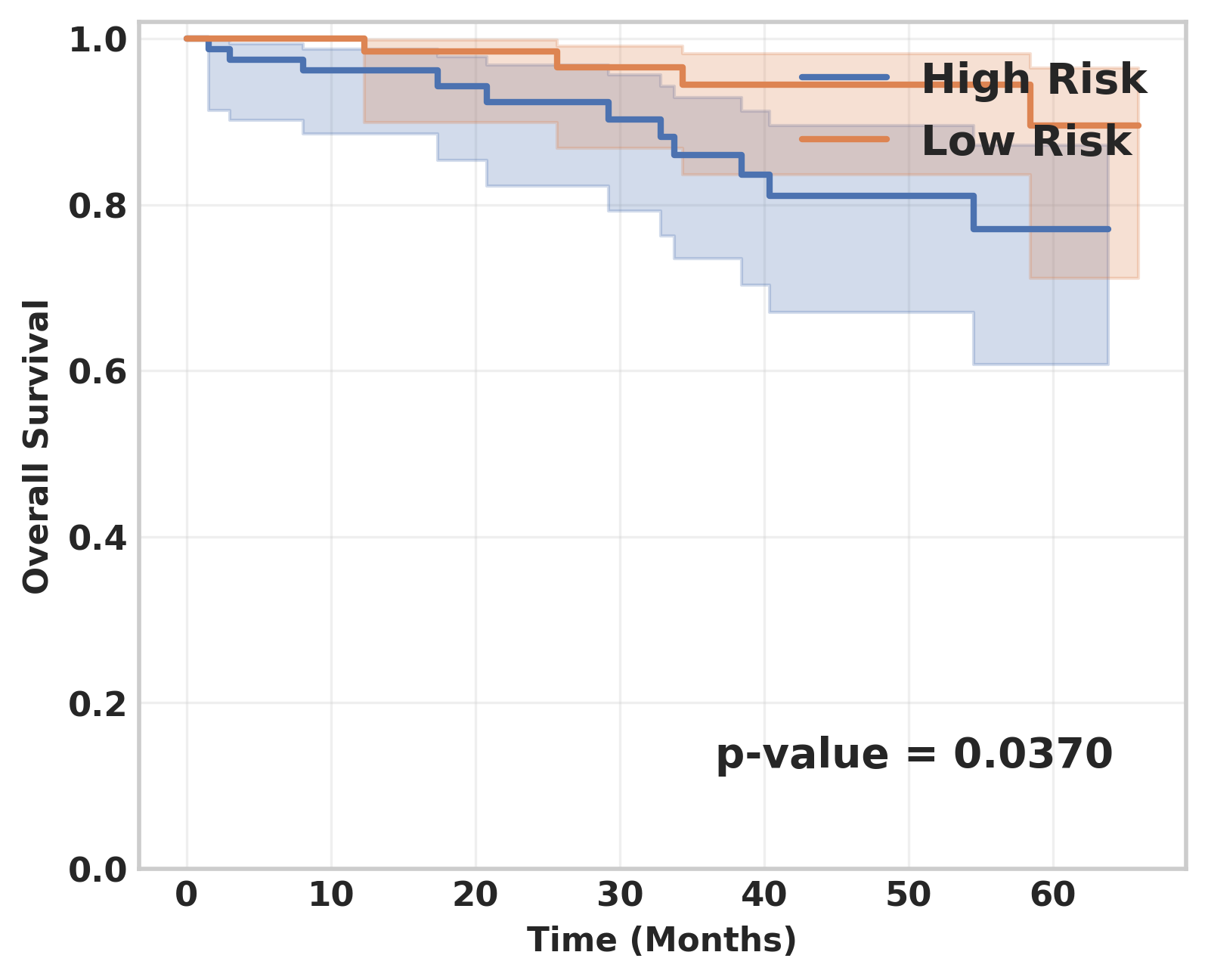}
    \caption{CPTAC-KIRC}
\end{subfigure}
\hfill
\begin{subfigure}{0.24\textwidth}
    \centering
    \includegraphics[width=\linewidth]{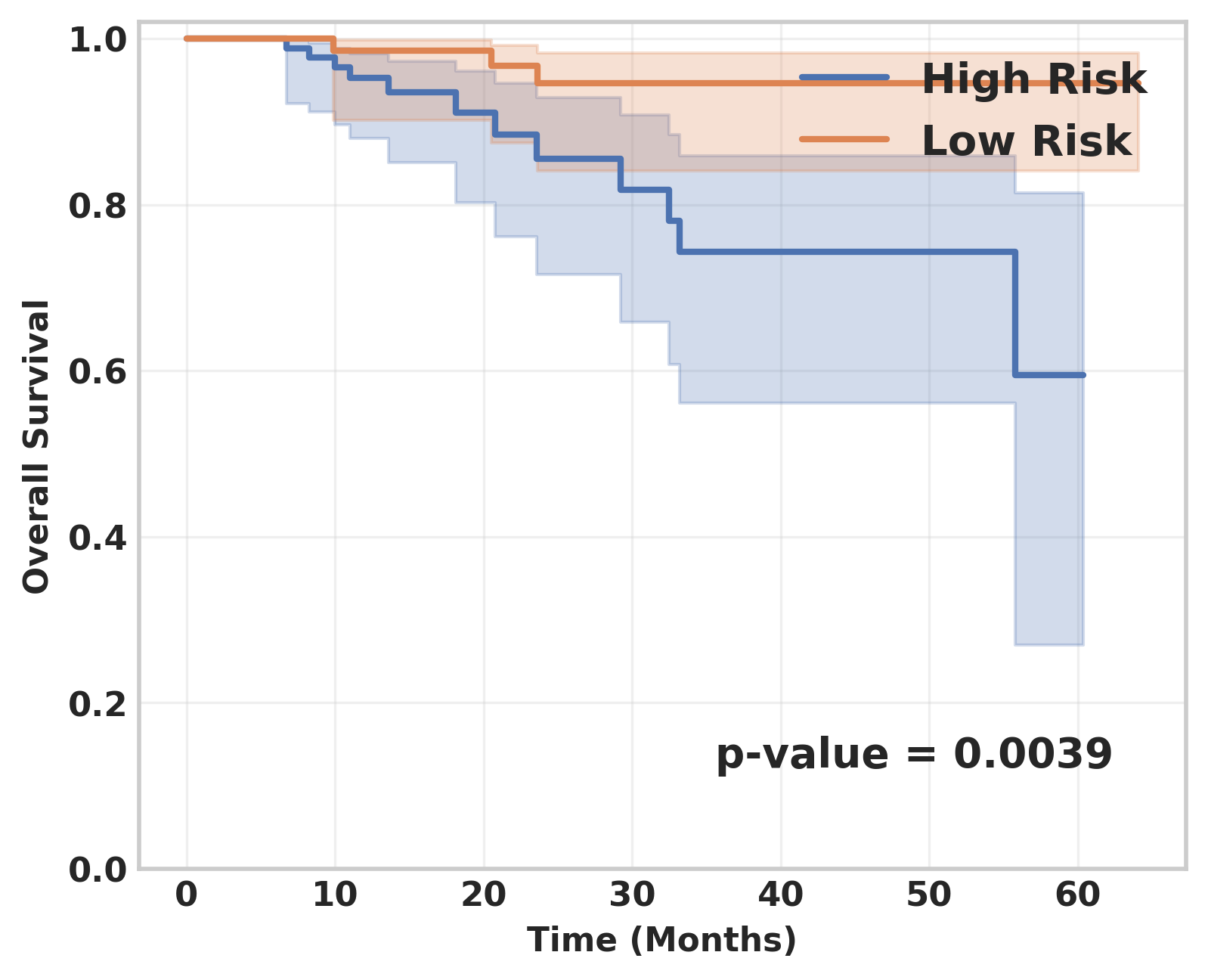}
    \caption{CPTAC-LUAD}
\end{subfigure}
\hfill
\begin{subfigure}{0.24\textwidth}
    \centering
    \includegraphics[width=\linewidth]{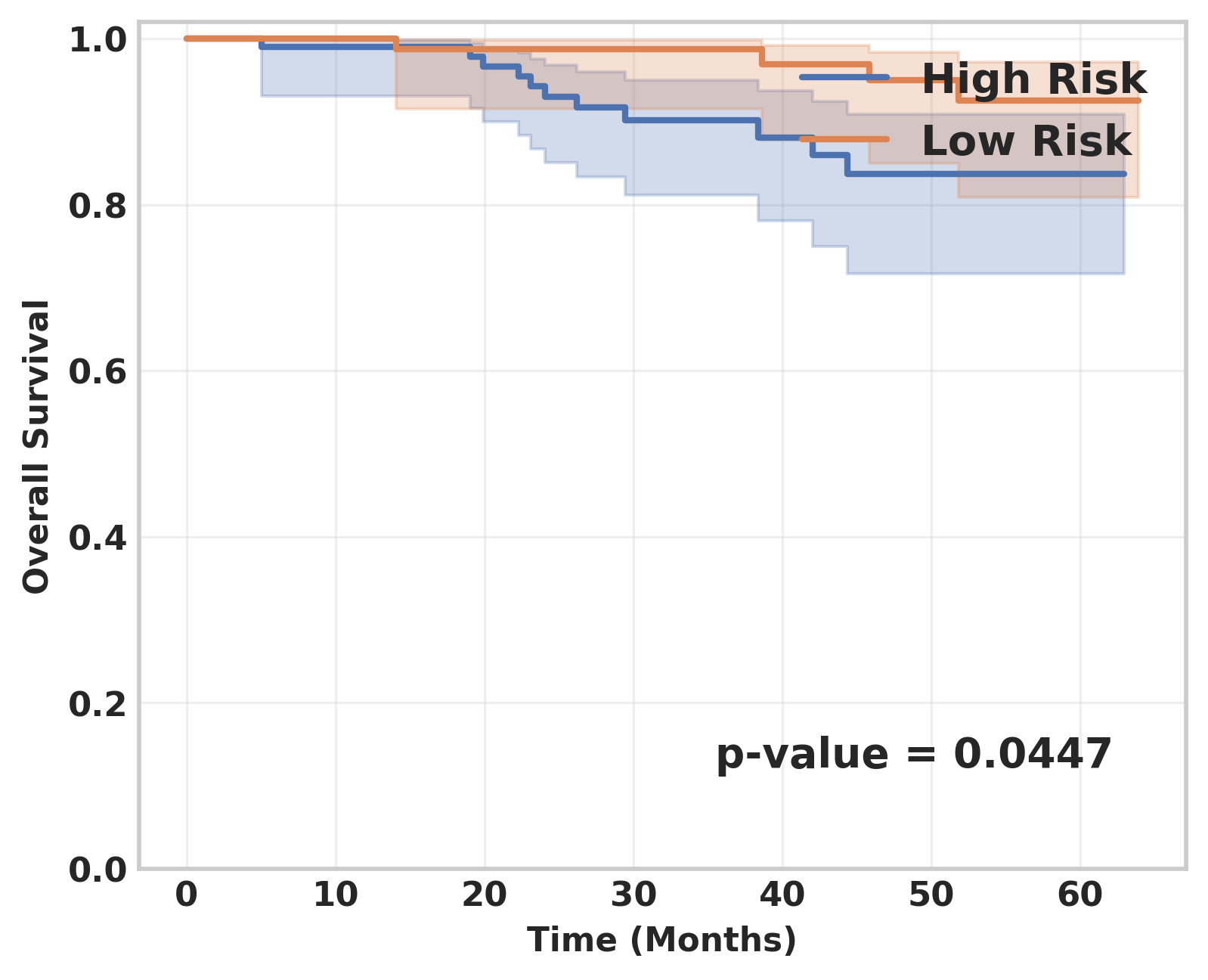}
    \caption{CPTAC-UCEC}
\end{subfigure}
\hfill
\begin{subfigure}{0.24\textwidth}
    \centering
    \includegraphics[width=\linewidth]{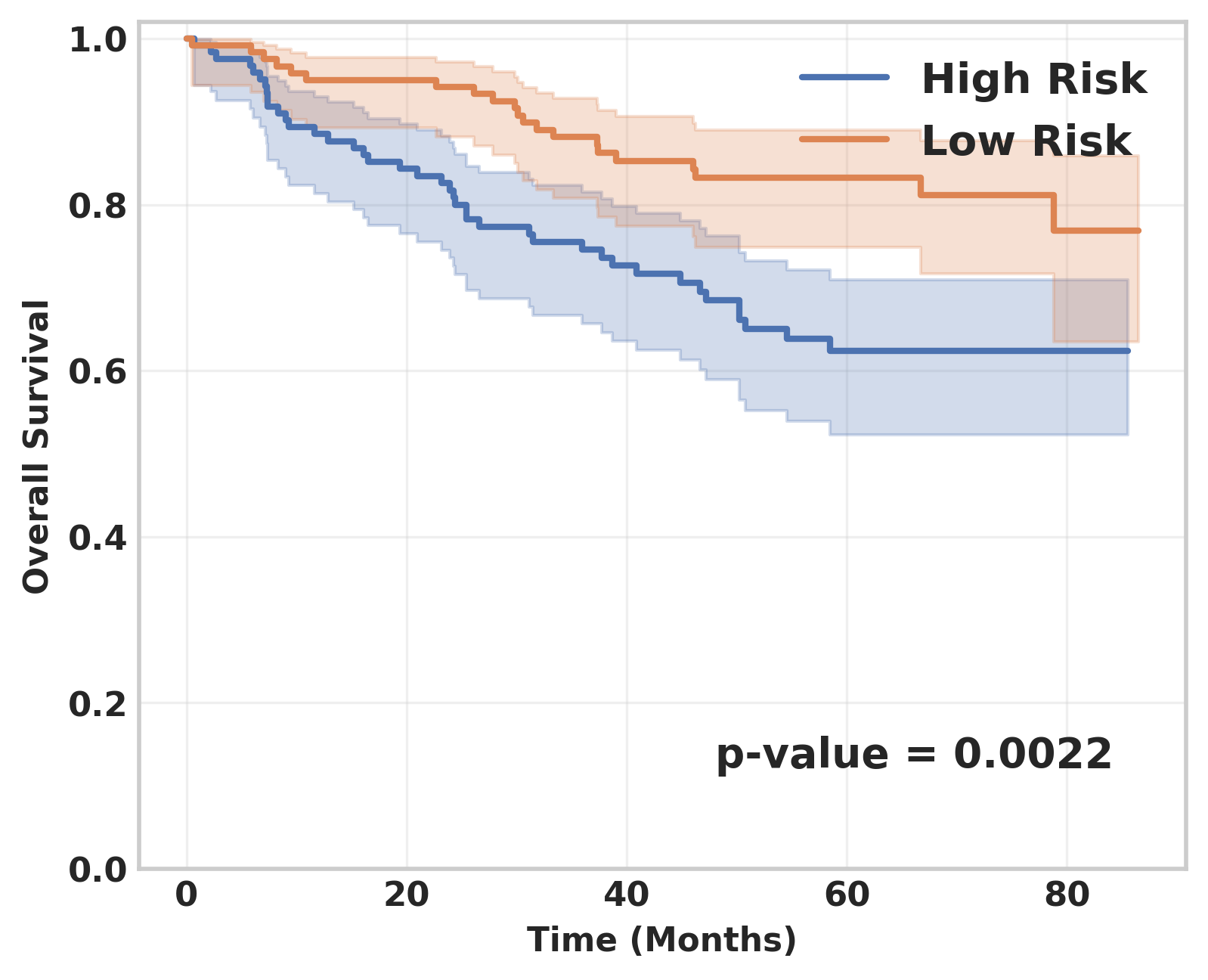}
    \caption{NLST-LUAD}
\end{subfigure}

\caption{
Kaplan–Meier risk stratification curves on external datasets 
(TCGA $\rightarrow$ CPTAC/NLST). 
High- and low-risk groups are separated by the median predicted risk.
}
\label{fig:km_external}
\end{figure*}

\noindent\textbf{Kaplan--Meier Risk Stratification.}
To assess whether SAEFS produces clinically meaningful risk predictions, we perform Kaplan-Meier survival analysis on each external cohort. Patients are stratified into high-risk and low-risk groups based on the median predicted risk score. As shown in Fig.~\ref{fig:km_external}, SAEFS achieves statistically significant stratification (log-rank test) between risk groups across all four external cohorts: CPTAC-KIRC ($p=0.0370$), CPTAC-LUAD ($p=0.0039$), CPTAC-UCEC ($p=0.0047$), and NLST-LUAD ($p=0.0022$). These results indicate that SAEFS produces consistent and clinically meaningful risk stratification even under substantial cross-center distribution shift.

\begin{figure*}[t!]
\centering

\begin{subfigure}{0.24\textwidth}
    \centering
    \includegraphics[width=\linewidth]{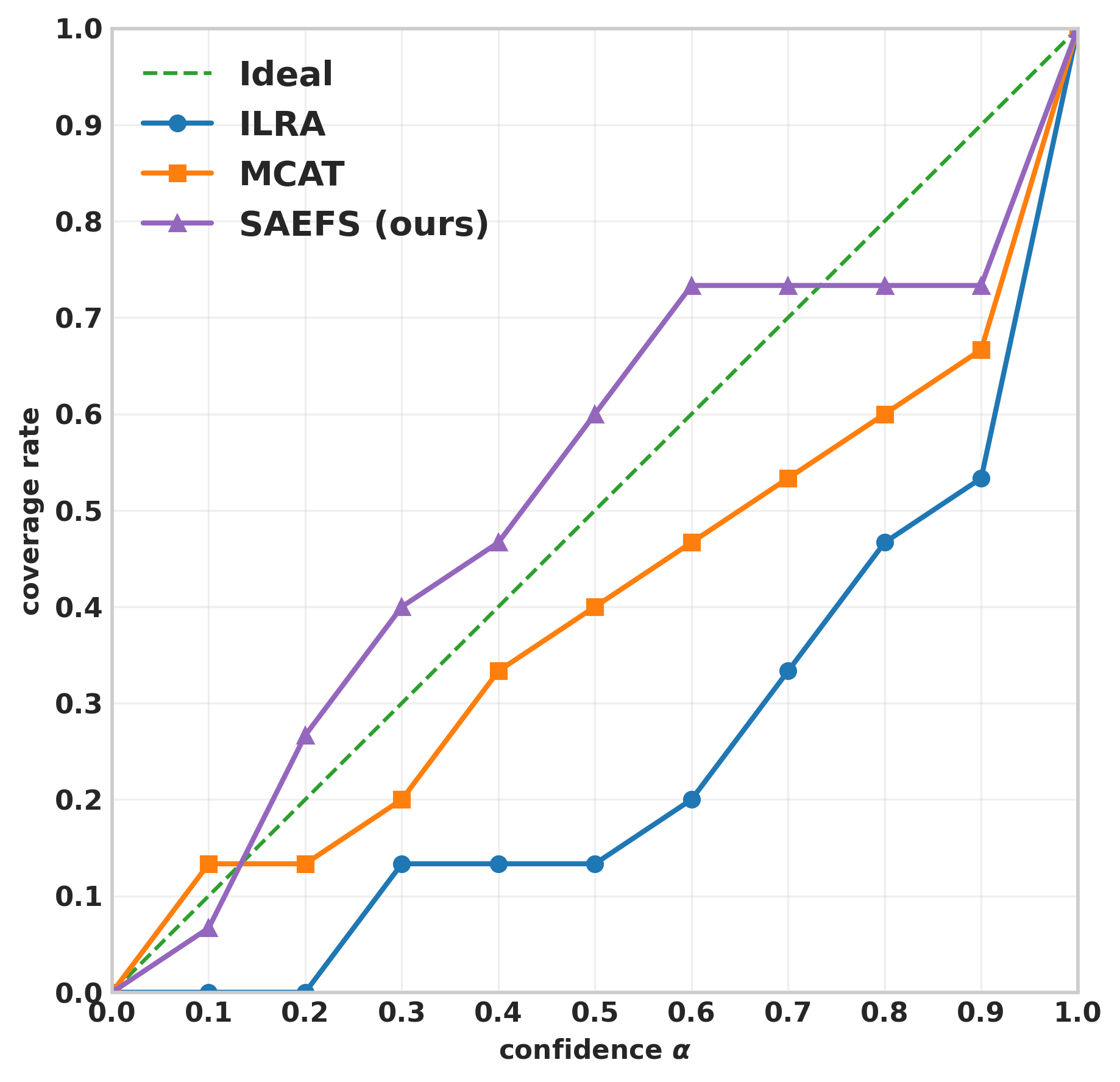}
    \caption{CPTAC-KIRC}
\end{subfigure}
\hfill
\begin{subfigure}{0.24\textwidth}
    \centering
    \includegraphics[width=\linewidth]{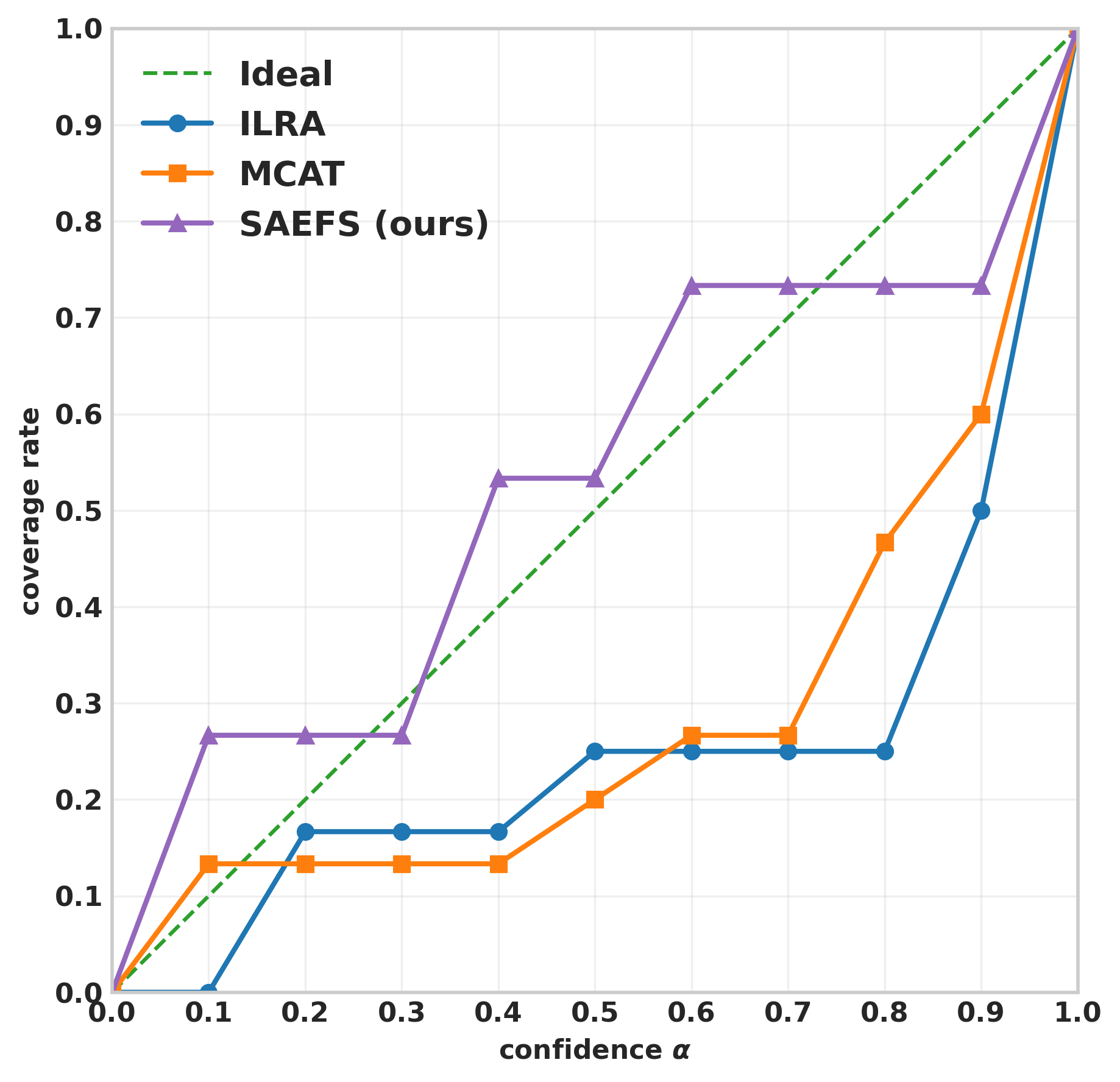}
    \caption{CPTAC-LUAD}
\end{subfigure}
\hfill
\begin{subfigure}{0.24\textwidth}
    \centering
    \includegraphics[width=\linewidth]{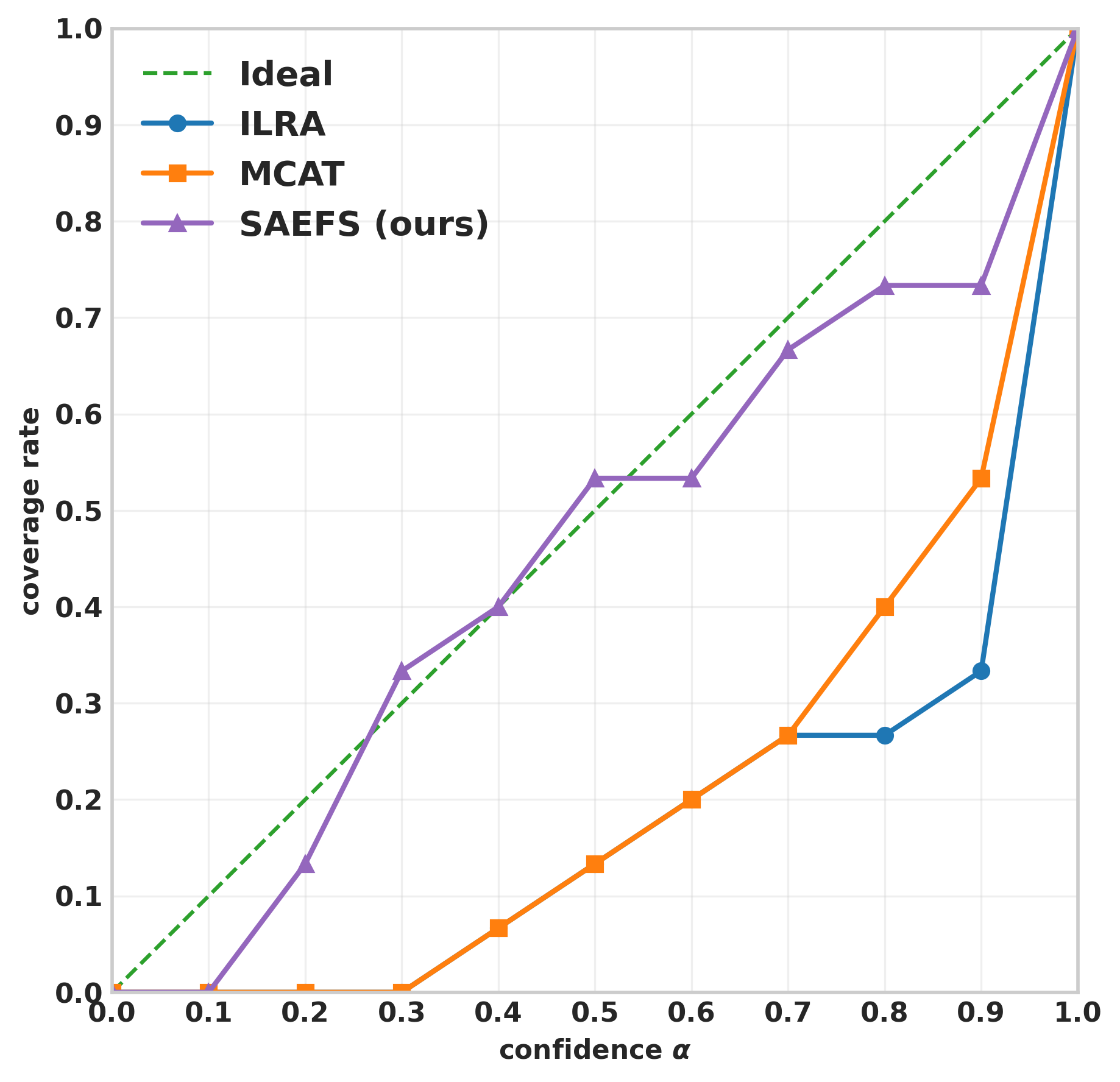}
    \caption{CPTAC-UCEC}
\end{subfigure}
\hfill
\begin{subfigure}{0.24\textwidth}
    \centering
    \includegraphics[width=\linewidth]{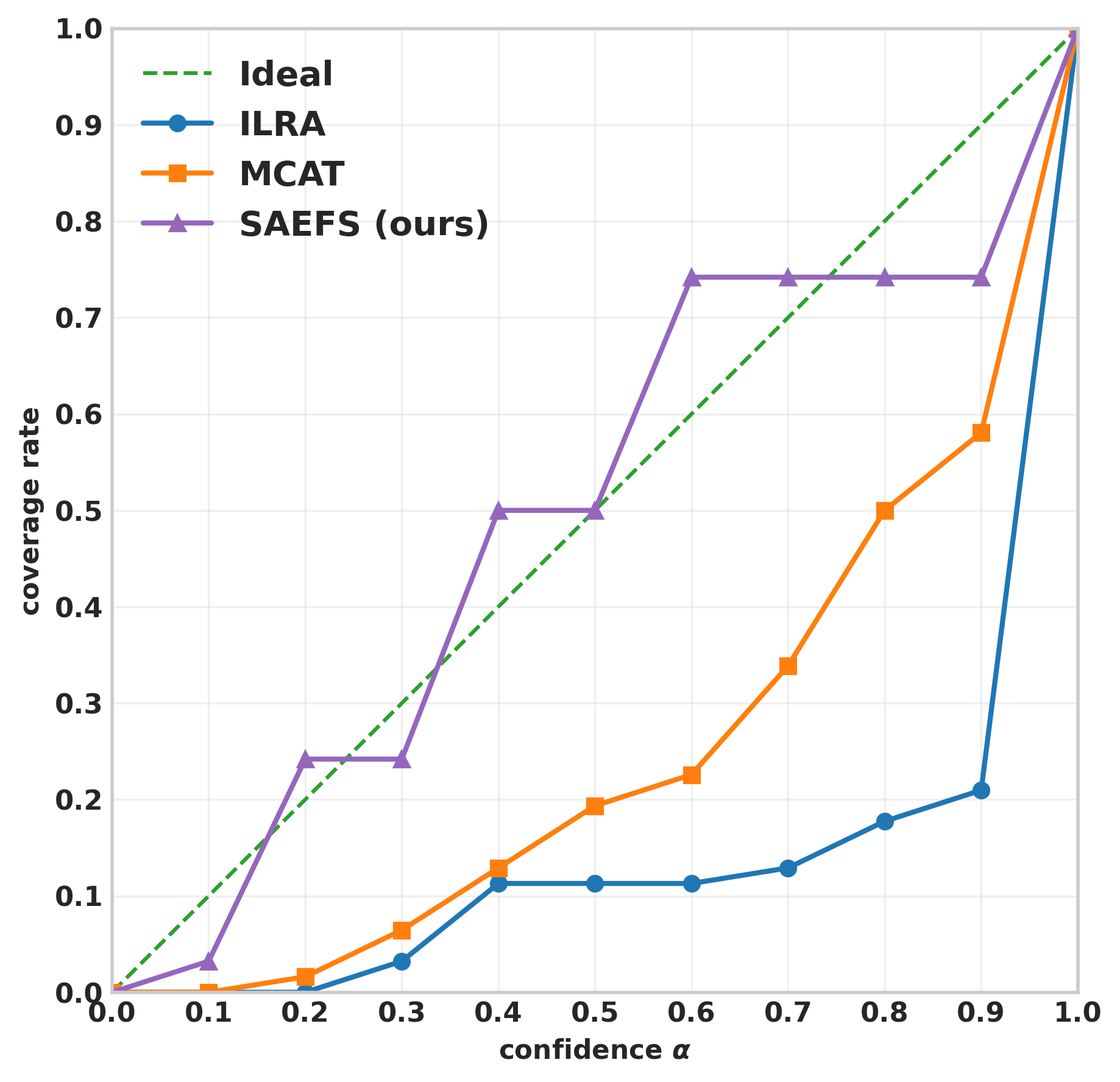}
    \caption{NLST-LUAD}
\end{subfigure}

\caption{
Calibration plots comparing predicted and observed survival probabilities on external cohorts
(TCGA $\rightarrow$ CPTAC/NLST). 
Curves closer to the diagonal indicate better calibration of survival risk predictions.
}
\label{fig:km_external_compare}
\end{figure*}

\noindent\textbf{Calibration Analysis.}
Beyond discrimination, reliable clinical deployment requires well-calibrated survival probabilities. Fig.~\ref{fig:km_external_compare} shows calibration curves comparing SAEFS with representative baselines across the four external cohorts. SAEFS consistently produces curves closer to the ideal diagonal, indicating improved calibration prediction under domain shift. 
Notably, although several multimodal baselines achieve slightly lower IBS and INBLL scores, their calibration curves deviate more noticeably from the diagonal line, suggesting less reliable probability estimates. The above observations indicate that SAEFS provides a better balance between predictive accuracy and probability calibration, attributable to the evidential modeling and cautious belief fusion mechanism.

\section{Conclusion}
\label{sec:conclusion}
In this work, we address the domain shift challenge in WSI survival analysis. We show that high-level pathological semantics are more domain-invariant than features captured by traditional MIL encoders. To operationalize this insight, we propose \textbf{SAEFS}, a survival framework that extracts semantics via template-structured VQA to anchor cross-center survival prediction in a semantic space, while employing cautious belief fusion to combine both visual and textual evidence. Zero-shot evaluations across four unseen multi-center cohorts show that SAEFS significantly outperforms state-of-the-art methods in prognostic accuracy and uncertainty calibration, yielding a more reliable and generalizable survival analysis framework. Future work will explore open-ended VLMs to derive finer semantic descriptions of rare morphological subtypes and investigate uncertainty-aware VLMs that provide semantic-level confidence for downstream diagnosis. By shifting from pixel-centric to clinical semantic-centric learning, our work advances a scalable paradigm for robust cross-center diagnostic systems.

\section*{Acknowledgement}
This research was supported by the National Medical Research Council (NMRC) Healthy and Meaningful Longevity--Cognition Grant (NICCOG2024-0028), the AI for Public Health Program at the Saw Swee Hock School of Public Health, National University of Singapore, and the Dame Julia Higgins Postdoc Collaborative Research Fund. This work was also supported by the NUS Guangzhou Research Translation and Innovation Institute (GRTII)-Affiliated PhD Scholarship Program.

\newpage
%
%
\bibliographystyle{splncs04}
\bibliography{main}
\end{document}